%% file: main.tex
\documentclass[11pt]{article}

\usepackage[final]{acl}

\usepackage{times}
\usepackage{latexsym}
\usepackage{amssymb}
\usepackage[T1]{fontenc}

\usepackage[utf8]{inputenc}

\usepackage{microtype}

\usepackage{inconsolata}
\usepackage{graphicx}
\usepackage{amsmath}
\usepackage{booktabs}
\usepackage{multirow}
\usepackage{xcolor,colortbl}
\usepackage{pifont}
\newcommand{\cmark}{\ding{51}}
\newcommand{\xmark}{\ding{55}}
\newcommand{\eg}{\textit{e.g.}}
\newcommand{\vs}{\textit{vs.}}
\usepackage{subcaption}
\usepackage{ulem}
\newcommand{\modelname}{LOVER}

\title{Long-to-Short Video Evidence Reasoning for \\ Grounded Question Answering}

\author{
  Kaiyan Chen \and
  Junbin Xiao\thanks{Corresponding authors.} \and
  Xun Yang$^*$ \\
  University of Science and Technology of China \\
  \texttt{chenky@mail.ustc.edu.cn, \{junbinxiao,xyang21\}@ustc.edu.cn}
}

\begin{document}
\maketitle

\input{secs/abstract}
\input{secs/intro}

\input{secs/related}
\input{secs/method}

\input{secs/experiment}
\input{secs/conclusion}

\bibliography{custom}

\clearpage
\appendix
\input{secs/appendix}
\end{document}

%% file: secs/abstract.tex
\begin{abstract}
We present \modelname, a \underline{L}ong to sh\underline{O}rt \underline{V}ideo \underline{E}vidence \underline{R}einforced model for grounded question answering (GQA). \modelname\ highlights three innovations over existing reinforcement-learning (RL) based video reasoning models: 
(1) \textbf{Long-to-short Video Evidence Curriculum Learning}, which organizes RL training according to evidence duration and progressively adapts the model from long-range grounding to short-term reasoning;
(2) \textbf{GQA Rewards}, which underscore the benefit of IoP reward over IoU for evidence spotting rather than strict temporal span overlap;
(3) \textbf{Adaptive Timestamp Rendering}, which adaptively renders timestamps onto video frames using background-aware position and color selection to enhance temporal observability.
The three designs are model-agnostic and reciprocal. They effectively improve QA, grounding, and grounded QA performance over different backbones. Notably, \modelname\ built on Time-R1 achieves new state-of-the-art (SOTA) results among open-source models on popular GQA benchmarks: NExT-GQA and ReXTime. Comprehensive ablation studies further validate the effectiveness of our three innovative components. 
\end{abstract}

%% file: secs/intro.tex
\section{Introduction}
\label{sec:intro}
Visual evidence grounded video question answering (GQA) \cite{xiao2024can,chen2024rextime,chen2025cg}, which requires models to answer the questions and temporally localize the supporting evidences, has emerged as an important task for reducing the hallucination of multimodal large language models (MLLMs) \cite{lu2025vited,gupta2025toga,bai2025qwen3} and improving their answer reliability to users. The task is challenging as it imposes substantially stronger demands on temporal reasoning and visual evidence spotting beyond question answering.

To tackle the challenges, existing approaches fall into either agentic reasoning \cite{min2024morevqa,zhang2024simple,xiao2025unleashing,liu2025videomind,dang2025mupa} or end-to-end learning \cite{wang2024grounded,meinardus2024chrono,zeng2025timesuite,lu2025vited,gupta2025toga}. The former seriously relies on pretrained capabilities and often with reduced efficiency because of multi-agent communication, while the latter often demands large-scale GQA-style data for instruction tuning.
Recent reinforcement learning (RL)-based post-training strategies \cite{wang2025time,yan2025videochat, bai2025qwen3} show promise for improved temporal grounding and question answering. Yet, they treat the two objectives separately, often leading to sub-optimal GQA performance which measures right answer for the right grounding.

In this paper, we take inspiration from RL post-training and propose \modelname, a \underline{L}ong to sh\underline{O}rt \underline{V}ideo \underline{E}vidence \underline{R}einforced model for GQA. 
\modelname\ highlights three innovative designs over exiting approaches: 
\textbf{(1) Long-to-Short Evidence Curriculum Learning}, which organizes RL training samples according to temporal evidence duration. The curriculum progressively transfers grounded reasoning capability from long-evidence scenarios to short ones, enabling the model to first learn easier temporal alignment before adapting to harder ones;
\textbf{(2) GQA Rewards}, which emphasize the advantage of \textit{Intersection over Prediction (IoP)} reward over IoU for temporal evidence spotting, rather than strict temporal span alignment, which we find often disturbs the optimization goal for QA;
\textbf{(3) Adaptive Timestamp Rendering}, which enhance timestamp perception of MLLMs by rendering the timestamps onto video frames with adaptive positions and colors in contrast to the foreground and background respectively. 

The three innovations respectively improves existing systems from the perspectives of training strategy, reward signal, and timestamp representation. Notably, all are architecture-independent and can be easily transferred to different backbones for consistent benefit.

We test \modelname\ on two standard video GQA benchmarks: NExT-GQA \cite{xiao2024can} and ReXTime \cite{chen2024rextime} with two backbones: Time-R1 \cite{wang2025time} and Qwen3-VL \cite{bai2025qwen3}. Experimental results show that \modelname\ consistently improves both temporal grounding and grounded QA, with GQA accuracy substantially exceeding baselines, and achieving the state-of-the-art (SOTA) among open-sourced models. 
Additionally, our extensive ablation studies and investigations demonstrate the effectiveness of three designs either independently or in cooperation, with enhanced benefits in answering questions whose temporal evidences are shorter in the videos.


Our contributions are summarized as follows:
\begin{itemize}
    \item We propose \textbf{\modelname} that highlights a suite of architecture-independent innovations for QA with improved vision trustability.
    \item We introduce a \textbf{Long-to-short Evidence Curriculum Learning} strategy that progressively reinforces evidence grounding from easy to hard samples.
    \item We follow the \textbf{Evidence Spotting Principle} and demonstrate IoP reward as a better alternative for GQA than IoU-based optimization.
    \item We propose \textbf{Adaptive Timestamp Rendering} that adaptively adjusts timestamp appearance across frames to enhance temporal observability while limiting visual interference.
\end{itemize}

%% file: secs/related.tex
\section{Related Work}

\subsection{Grounded Video Question Answering}
Recent advances in MLLMs have substantially improved video question answering through large-scale pretraining and instruction tuning \cite{maaz2024video,li2024llava,lin2024vila,zhang2024llava,wang2025internvl3,bai2025qwen3}. GQA further requires models to explicitly localize the temporal evidence supporting the predicted answer, making temporal grounding an intrinsic component of answer reasoning rather than an additional prediction task.
Existing methods improve grounded video reasoning through agentic or tool-augmented reasoning \cite{min2024morevqa, zhang2024simple, xiao2025unleashing, shi2025enhancing}, or end-to-end learning with refined temporal representations \cite{lu2025vited,gupta2025toga,jung2025consistency,zeng2025factorized,wu2025number}, or RL-based post-training \cite{wang2025time,feng2025video,yan2025videochat}.

Despite recent progress, existing methods largely ignore the imbalanced difficulty of temporal evidence in videos. Long evidence is often redundant and easier to localize, whereas short evidence is sparse, highly sensitive to localization errors, and harder to identify. Existing approaches optimize all samples uniformly, leading to sub-optimal performance. 
In addition, most methods decouple grounding from QA and rely on IoU-based objectives, which prioritize precise boundary alignment over evidence spotting, though a few key frames are often sufficient for answering questions. This interferes with QA rather than enhances it, due to heavy boundary annotation noises and significantly increased optimization difficulty. 
Moreover, most systems perform implicit temporal inference from timestamp embeddings, lacking explicitly visible time cues for temporal reasoning. \modelname\ addresses the above limitations by introducing 1) stage-wise curriculum learning according to evidence duration, 2) an additional IoP reward tailored for GQA, and 3) an adaptive timestamp rendering strategy to enhance visible time cues.

\subsection{Curriculum Learning in MLLMs}
Curriculum Learning (CL) progressively organizes training samples according to task difficulty, enabling more stable optimization and improved generalization \cite{bengio2009curriculum,wang2021survey}. Such strategies have shown effectiveness in multimodal reasoning and reinforcement learning \cite{dong2025videotg,jin2025videocurl}. Existing MLLM curricula mainly rely on data filtering \cite{gadre2023datacomp,lee2024concept,ma2025mllm, wang2025sota}, staged training, or model-specific difficulty signals such as loss, confidence, or uncertainty \cite{li2024answering}. However, these model-internal signals do not accurately reflect the intrinsic difficulty structure of GQA.

Different from prior curricula that rely on indirect difficulty proxies, our long-to-short evidence curriculum uses ground-truth interval duration as a direct and intrinsic measure of task difficulty, progressively transferring grounded reasoning capability from long video evidences to short ones without requiring any model-internal signals.
\begin{figure*}[t!]
    \centering
    \includegraphics[width=\textwidth]{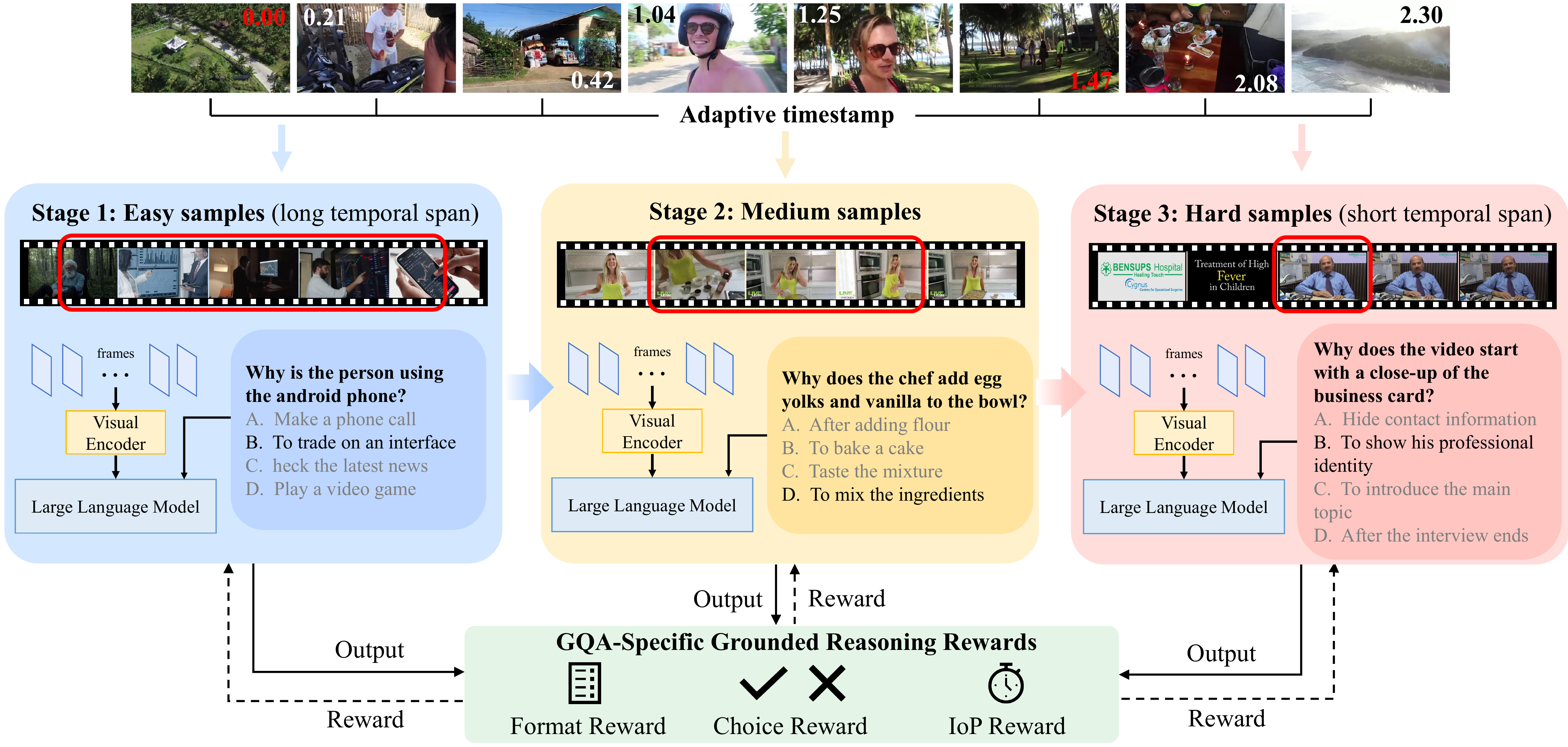}
    \caption{Overview of \modelname\ method. Top: Video frames of adaptively rendered timestamps. Middle: Training sample organization according to evidence duration. Bottom: GQA rewards composing IoP, answer choice correctness, and format rewards.}
    \label{fig:curriculum}
    \vspace{-3mm}
\end{figure*}

\subsection{Timestamp Representation in VideoLLMs}
Most existing methods capture temporal information using temporal position embeddings \cite{ren2024timechat,lu2025vited,gupta2025toga,meinardus2024chrono}. Recently, NumPro \cite{wu2025number} showed that rendering timestamps on frames can improve temporal reasoning. However, it uses fixed timestamp positions and font colors, which may occlude foreground content or blend into the background. Moreover, timestamp rendering has not been explored in RL-based temporal evidence reasoning. In contrast, our adaptive timestamp rendering dynamically adjusts timestamp positions and font colors according to frame content during RL post-training, yielding more effective temporal prompting and better performance.

%% file: secs/method.tex
\section{Method}
\subsection{Problem and Solution Overview}
Given a video $V$, a question $Q$, and candidate answers $\mathcal{A}$, 
GQA aims to predict both the answer $A^{*}$ and the supporting 
temporal interval $\mathbf{g}=(t_s, t_e)$. Following recent RL-based frameworks, the model autoregressively generates:
\begin{equation}
Y = \{ \hat{t}_s,\ \hat{t}_e,\ \hat{A},\ \hat{R} \}
\end{equation}
where $\hat{t}_s, \hat{t}_e$ denote predicted temporal boundaries, $\hat{A}$ is the predicted answer, and $\hat{R}$ denotes the reasoning process.

To solve GQA, we build upon existing MLLMs with RL post-training and propose \modelname\ (Fig.~\ref{fig:curriculum}), featuring three key innovations for better performance. First, we introduce \textbf{Long-to-Short Evidence Curriculum Learning}, which organizes RL samples by evidence duration and progressively transfers grounded reasoning ability from long, easier evidences to short, challenging ones. Second, we design \textbf{GQA Rewards}, highlighting the advantage of IoP over IoU by emphasizing evidence spotting rather than strict temporal boundary alignment. Third, we propose \textbf{Adaptive Timestamp Rendering}, which improves timestamp perception by adaptively rendering timestamps with content-aware positions and colors on video frames. All three components are architecture-agnostic and can be seamlessly applied to different backbones for consistent gains. Details are provided in the following sections. 

\subsection{Long-to-Short Evidence Curriculum}
To progressively transfer grounded reasoning ability from long to short video evidence, we introduce a \textbf{Long-to-Short Evidence Curriculum}. Given a temporal interval $\mathbf{g}=(t_s,t_e)$ with duration $d=t_e-t_s$, we use $d$ as an intrinsic measure of task difficulty: long intervals contain dense and redundant evidence with high localization tolerance, whereas short intervals provide sparse evidence that requires precise grounding. Importantly, this difficulty measure is derived directly from ground-truth annotations, without relying on model-dependent signals such as loss or confidence.

Training samples are divided into three stages according to descending evidence duration, using a 3:2:2 allocation ratio to emphasize long-evidence samples in early training. The model first learns coarse temporal alignment on long-evidence samples, and then progressively adapts to fine-grained grounding on shorter evidence.

\subsection{GQA Rewards}
Existing grounding and QA methods treat the two objectives separately and optimize IoU for grounding. IoU overly emphasizes precise time span alignment rather than evidence spotting, even though several key frames are sufficient to answer most questions (\eg, a single frame can answer most recognition questions). The noising temporal boundary annotations and increased task difficulty often lead to unintended-degree of performance improvements. Exact overlap is therefore a misaligned objective for GQA.

Under the above evidence spotting principle, we instead introduce IoP 
(Intersection over Prediction) as a better aligned reward for GQA. IoP 
directly incentivizes evidence precision rather than time-span overlap:
\begin{equation}
\text{IoP}(\hat{\mathbf{g}}, \mathbf{g}) = 
\frac{|\hat{\mathbf{g}} \cap \mathbf{g}|}{|\hat{\mathbf{g}}|}
\end{equation}
where $\hat{\mathbf{g}}$ and $\mathbf{g}$ denote the predicted and ground-truth time intervals, respectively. If a predicted interval $\hat{\mathbf{g}}$
falls into ground-truth $\mathbf{g}$, it will get full reward for grounding precision.

The final GQA rewards combine grounded reasoning quality, answer correctness, and 
output format:
\begin{equation}
\mathcal{R} = \lambda_1 \mathcal{R}_{\text{IoP}} + \lambda_2 
\mathcal{R}_{\text{ans}} + \lambda_3 \mathcal{R}_{\text{format}}
\end{equation}
where $\mathcal{R}_{\text{IoP}}$ serves as the grounded reasoning reward 
evaluating evidence spotting for GQA, $\mathcal{R}_{\text{ans}}$ evaluates 
answer correctness, and $\mathcal{R}_{\text{format}}$ encourages structured 
outputs.

\begin{figure}[t]
    \centering
    \includegraphics[width=0.48\textwidth]{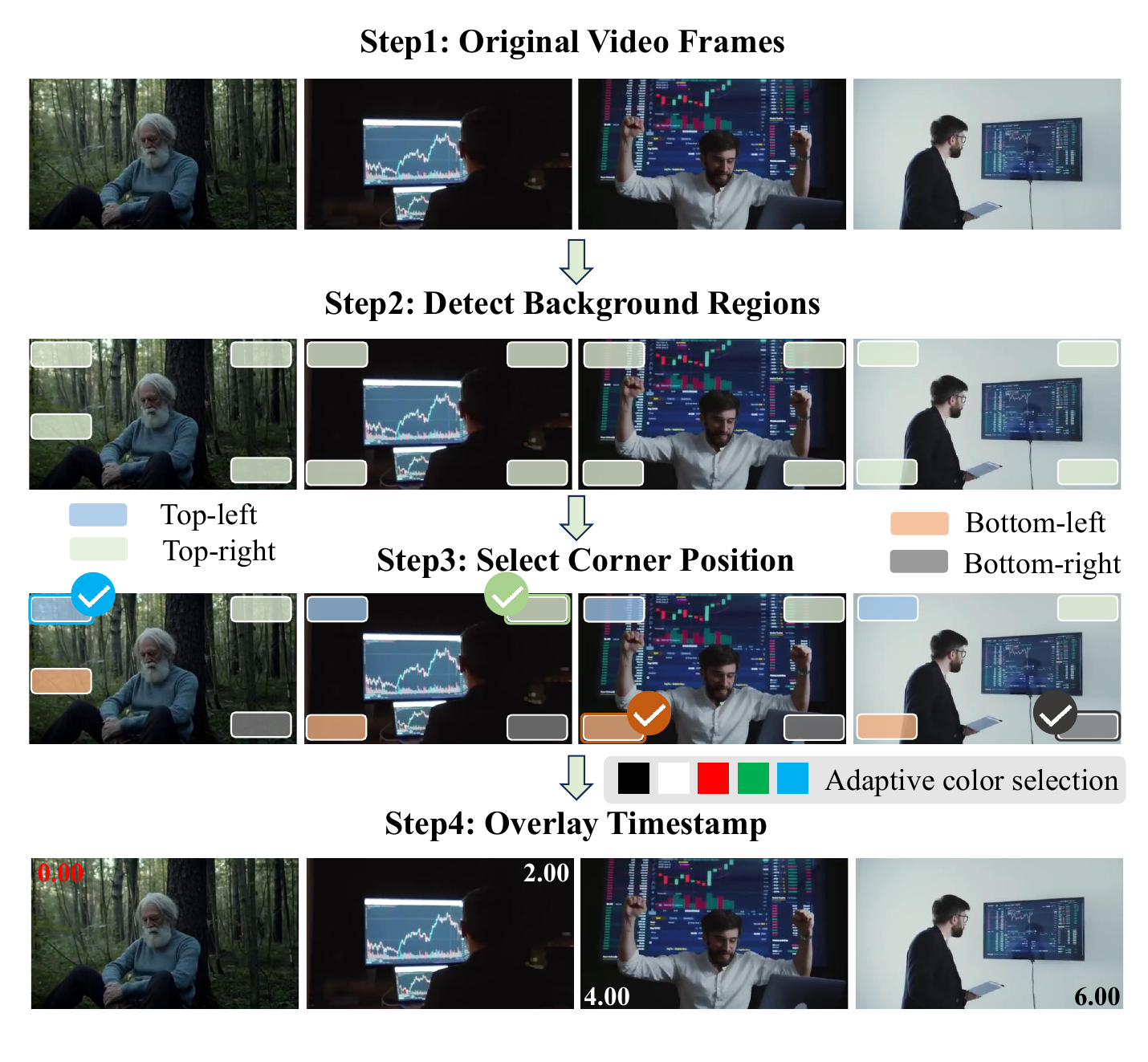}
    \caption{Position- and color-adaptive time rendering. This figure illustrates the four-step pipeline: extracting original frames, detecting background regions, selecting low-interference corner positions, and overlaying timestamps with adaptively chosen font colors.}
    \label{fig:timenum}
    \vspace{-3mm}
\end{figure}

\subsection{Adaptive Timestamp Rendering}
Recent approaches have shown the effectiveness of visual rendering for improved grounding \cite{yao2024cpt,wu2025number}. Thus, we improve the temporal perception capability of MLLMs by explicitly rendering timestamps onto sampled frames:
\begin{equation}
f_i' = \Phi(f_i,\ t_i)
\end{equation}
where $f_i$ is the sampled frame at time $t_i$, and $\Phi(\cdot)$ denotes timestamp rendering. However, fixed-style rendering may occlude visual content or blend into the background, weakening the visual prompting effect. We therefore introduce \textbf{adaptive timestamp rendering} (Fig.~\ref{fig:timenum}), which dynamically places timestamps in low-interference boundary regions and adaptively selects font colors based on local contrast for better readability (Detailed implementations are presented in the Appendix). This design requires no architectural modification, while providing explicit temporal cues to reduce ambiguity.

\subsection{Overall Optimization}
\modelname\ jointly integrates the long-to-short evidence curriculum, GQA Rewards, and background-aware adaptive timestamp rendering into 
a unified RL framework. The overall objective is:
\begin{equation}
\max_{\theta}\ \mathbb{E}_{Y \sim \pi_{\theta}}\left[\mathcal{R}(Y)\right]
\end{equation}
where $\pi_{\theta}$ denotes the policy model and $\mathcal{R}(Y)$ denotes the combined evidence-aware reward. By jointly addressing evidence sparsity across training strategy, reward design, and input observability, \modelname\ enables stable and effective grounded video reasoning under reinforcement learning.

%% file: secs/experiment.tex
\section{Experiments}

\begin{table*}[t]
\centering
\scriptsize
\setlength{\tabcolsep}{3.5pt}

\begin{subtable}[t]{0.49\textwidth}
\centering
\renewcommand{\arraystretch}{0.78}
\begin{tabular}{lcccccc}
\toprule
\textbf{Model} & \textbf{Size} & \textbf{mIoP} & \textbf{mIoU} & \textbf{IoP@.5} & \textbf{IoU@.5} & \textbf{GQA} \\
\midrule
\multicolumn{7}{l}{\cellcolor[rgb]{0.898,0.898,0.898}\textit{Agentic}} \\
LangRepo & 7B & 31.3 & 18.5 & 28.7 & 12.2 & 17.1 \\
LLoVi (GPT-4) & - & 37.3 & 20.0 & 36.9 & 15.3 & 24.3 \\
VideoMind & 7B & 39.0 & 31.4 & 35.3 & 25.8 & 28.2 \\
DeVi (Gemini-2) & - & 39.7 & 23.6 & 38.9 & 19.5 & 28.9 \\
\midrule
\multicolumn{7}{l}{\cellcolor[rgb]{0.898,0.898,0.898}\textit{End-to-End}} \\
Temp(CLIP) & 200M & 25.7 & 12.1 & 25.5 & 8.9 & 16.0 \\
SeViLA & 4B & 29.5 & 21.7 & 22.9 & 13.8 & 16.6 \\
FrozenBiLM & 890M & 24.2 & 9.6 & 23.7 & 6.1 & 17.5 \\
TOGA & 7B & 40.5 & 24.4 & 40.6 & 21.1 & 24.6 \\
VideoChat-TPO & 7B & 35.6 & 27.7 & 32.8 & 23.4 & 25.5 \\
Grounded-LLM & 7B & 34.5 & 21.1 & 34.4 & 18.0 & 26.7 \\
VideoStream & 8.3B & 32.2 & 19.3 & 31.0 & 13.3 & 17.8 \\
Qwen3-VL* & 4B & 32.1 & 19.4 & 27.3 & 12.5 & 22.1 \\
Time-R1* & 7B & 36.0 & 30.7 & 31.9 & 25.3 & 26.1 \\
\rowcolor[rgb]{0.949,0.996,0.949}
\modelname (Qwen3-VL) & 4B & 36.2 & 20.4 & 35.3& 16.7 & 27.5 \\
\rowcolor[rgb]{0.949,0.996,0.949}
\modelname (Time-R1) & 7B & \textbf{42.7} & \textbf{33.5} & \textbf{41.1} & \textbf{29.8} & \textbf{32.7} \\
\bottomrule
\end{tabular}
\caption{NExT-GQA}
\label{tab:nextgqa}
\end{subtable}
\hfill
\begin{subtable}[t]{0.49\textwidth}
\centering
\begin{tabular}{lcccccc}
\toprule
\textbf{Model} & \textbf{Size} & \textbf{mIoU} & \textbf{IoU@.3} & \textbf{IoU@.5} & \textbf{Acc} & \textbf{GQA} \\
\midrule
\multicolumn{7}{l}{\cellcolor[rgb]{0.898,0.898,0.898}\textit{Closed-source}} \\
Gemini-1.5-Pro & - & 28.43 & 35.67 & 25.00 & 68.00 & 13.3 \\
Claude-3-Opus & - & 23.61 & 30.67 & 17.67 & 68.67 & 13.67 \\
GPT-4V & - & 26.74 & 33.33 & 22.00 & 63.33 & 16.67 \\
Reka-Core & - & 27.95 & 36.33 & 24.00 & 59.67 & 17.00 \\
GPT-4o & - & \textbf{36.28} & \textbf{45.33} & \textbf{34.00} & \textbf{73.67} & \textbf{28.67} \\
\midrule
\multicolumn{7}{l}{\cellcolor[rgb]{0.898,0.898,0.898}\textit{Open-source}} \\
VTimeLLM & 7B & 20.14 & 28.84 & 17.41 & 36.16 & - \\
TimeChat & 7B & 11.65 & 14.42 & 7.61 & 40.04 & - \\
LITA & 13B & 21.49 & 29.49 & 16.29 & 34.44 & - \\
Qwen3-VL* & 4B & 20.97 & 26.23 & 20.11 & 65.19 & 13.23 \\
\rowcolor[rgb]{0.949,0.996,0.949}
\modelname\ (Qwen3-VL) & 4B & 22.02 & 27.37 & 21.48 & 65.64 & 14.81 \\
Time-R1* & 7B & 28.40 & 39.53 & 27.37 & 71.18 & 21.46 \\
\rowcolor[rgb]{0.949,0.996,0.949}
\modelname\ (Time-R1) & 7B & \underline{29.39} & \underline{41.64} & \underline{27.37} & \underline{72.93} & \underline{21.87} \\
\bottomrule
\end{tabular}
\caption{ReXTime}
\label{tab:rextime}
\end{subtable}

\vspace{-2mm}
\caption{Results on NExT-GQA and ReXTime benchmarks.}
\label{tab:compact_results}
\vspace{-3mm}
\end{table*}

\subsection{Experimental Setup}

\noindent\textbf{Benchmarks.}
We evaluate \modelname\ on two standard GQA benchmarks.
\textit{NExT-GQA}~\cite{xiao2024can} contains 990 videos and 5,553 QAs, with an average video length of $\sim$40 seconds. Each video is associated with multiple questions involving causal reasoning (\textit{why}/\textit{how}) and temporal reasoning  \textit{when}/\textit{before}/\textit{after}), and the supporting evidence for different questions may overlap in time. Models are required to jointly predict answers and localize the supporting temporal interval. 
\textit{ReXTime}~\cite{chen2024rextime} includes 921 validation samples and 2,143 test samples. The videos are long, with an average length of $\sim$3 minutes. ReXTime focuses on a more challenging scenario where the question and its answer evidence reside in different temporal segments of the video, necessitating cross-segment causal and temporal reasoning.  For both datasets, we follow standard protocols for evaluation. Specially, Acc@GQA measures the percentages of correct answer with correct grounding (IoP$\ge$0.5 for NExT-GQA or IoU$\ge$0.5 for ReXTime unless otherwise specified).We also include an additional evaluation benchmark, with its details provided in Appendix~\ref{CG-Bench}.

\noindent\textbf{Training Data.}
We start from Time-R1's 2,500 temporally grounded video descriptions and reformulate them into GQA-style supervision via DeepSeek-assisted rewriting: converting each sample into a multiple-choice QA pair along with its corresponding temporal evidence interval.we partition the training set into three difficulty stages according to ground-truth evidence duration. The partition thresholds are set to 7s and 15s for short (0-7s), middle (7-15s) and long ($\ge$15s) durations, which are determined following a training data ratio of 3:2:2.

\noindent\textbf{Implementation Details.}
We instantiate \modelname\ with two recent backbones: Time-R1-7B \cite{wang2025time} and Qwen3-VL-4B~\cite{bai2025qwen3}. To test the reliability of LLM in data rewriting, we manually selected and examined 500 random samples, and achieved a pass rate of 96.8\%. For details of the analysis on the quality of the training data, please refer to Appendix~\ref{Quality_Verification}.

\subsection{Comparison with SOTAs}
Table~\ref{tab:nextgqa} presents results on NExT-GQA. \modelname\ (Time-R1) consistently outperforms all methods across different metrics. Moreover, the gains over baselines (Qwen3-VL or Time-R1) are pronounced for both temporal grounding and grounded QA. Table~\ref{tab:rextime} further confirm \modelname's steady improvements over baselines, and being the best among the open-source models. Yet, the shrunken improvements could be due to
a deviated challenge that the evidences for question and answer are speared at different places of a video.
Interestingly, both Tables~\ref{tab:nextgqa} and \ref{tab:rextime} show that our method effectively improves IoU-based metrics even without IoU reward, demonstrating IoP as a reasonable or even superior alternative.
\begin{table*}[t!]
\centering
\small
\scalebox{0.73}{
\begin{tabular}{ll ccc ccc ccc}
\toprule
\multirow{2}{*}{Dataset} & \multirow{2}{*}{Metric}
& \multicolumn{3}{c}{Long}
& \multicolumn{3}{c}{Medium}
& \multicolumn{3}{c}{Short} \\
\cmidrule(lr){3-5} \cmidrule(lr){6-8} \cmidrule(lr){9-11}
& & Time-R1 & \shortstack{LOVER(w/o TS)} & LOVER 
& Time-R1 & \shortstack{LOVER(w/o TS)} & LOVER 
& Time-R1 & \shortstack{LOVER(w/o TS)} & LOVER \\
\midrule
\multirow{4}{*}{NExT-GQA}
& mIoP    & 76.06 & 80.48 & 82.06 & 52.49 & 55.79 & 61.13 & 27.06 & 29.77 & 33.23 \\
& IoP@0.5 & 80.56 & 85.17 & 85.97 & 57.60 & 60.70 & 65.70 & 19.59 & 23.31 & 29.48 \\
& Acc@QA  & 82.36 & 80.16 & 80.36 & 80.80 & 80.50 & 79.80 & 77.21 & 75.41 & 75.41 \\
& GQA(IoP$\ge$0.5) & 67.13 & 70.14 & \textbf{70.14} & 46.70 & 50.50 & \textbf{52.90} & 15.91 & 18.20 & \textbf{23.09} \\
\midrule
\multirow{4}{*}{ReXTime}
& mIoP    & 53.19 & 62.47 & 60.50 & 39.11 & 39.94 & 46.44 & 25.73 & 25.32 & 29.93 \\
& IoP@0.5 & 57.42 & 64.37 & 63.86 & 45.38 & 43.85 & 51.15 & 20.24 & 21.67 & 26.19 \\
& Acc@QA  & 74.06 & 73.35 & 75.49 & 72.31 & 68.46 & 77.69 & 60.71 & 60.71 & 60.71 \\
& GQA(IoP$\ge$0.5) & 45.97 & 50.59 & \textbf{50.98} & 34.62 & 37.31 & \textbf{39.23} & 14.29 & 16.10 & \textbf{19.64} \\
\midrule
\multirow{2}{*}{$\Delta$ GQA(IoP$\ge$0.5)}
& NExT-GQA & \multicolumn{3}{c}{+3.01 / +0.00} & \multicolumn{3}{c}{+6.20 / +2.40} & \multicolumn{3}{c}{+7.18 / +4.89} \\
& ReXTime  & \multicolumn{3}{c}{+5.01 / $+$0.39} & \multicolumn{3}{c}{$+$4.61 / +1.92} & \multicolumn{3}{c}{$+$5.35 / +3.54} \\
\bottomrule
\end{tabular}
}
\caption{Performance across durations. \modelname's gains grow progressively as evidence duration decreases, with the largest improvements on short-evidence samples. Also, adaptive timestamp rendering helps more on samples of short evidences. xx / xx: (\modelname\ \vs~ Time-R1) ~/~ (\modelname\ \vs~ its variant without timestamp (TS) rendering).}
\label{tab:duration_analysis}
\end{table*}

\subsection{Analysis Across Evidence Durations}
Table~\ref{tab:duration_analysis} breaks down performance across three evidence duration groups — long, medium, and short. \modelname\
consistently outperforms Time-R1 in Acc@GQA across all groups, and the superiority gets enhanced progressively as evidence duration decreases. The most substantial gains are observed on short-evidence samples. This pattern directly validates the design motivation of the Long-to-Short Evidence Curriculum: by progressively exposing the model to harder short-evidence samples after establishing coarse alignment on long-evidence ones, the model acquires more precise grounding capability where it is most needed.

Again, adaptive timestamp rendering contributes across all evidence-duration groups, with its benefit particularly evident on short-evidence samples where explicit temporal cues are most critical for disambiguating sparse visual evidence. Notably, Acc@QA remains largely stable across configurations, indicating that the grounding improvements stem from better temporal localization rather than from changes in answer generation.

\begin{table}[t]
\centering
\small

\begin{tabular}{lcccccc}
\toprule

\multirow{2}{*}{\textbf{Methods}}
& \multirow{2}{*}{\textbf{IoP}}
& \multirow{2}{*}{\textbf{TS}}
& \multirow{2}{*}{\textbf{CL}}
& \multicolumn{2}{c}{\textbf{GQA(IoP$\ge$.5)}} \\

\cmidrule(lr){5-6} 
& & & & \textbf{ReXT} & \textbf{NExT} \\
\midrule

Time-R1 & \xmark & \xmark & \xmark & 38.04 & 26.06 \\

Baseline & \xmark & \xmark & \xmark & 40.37 & 27.10 \\

\midrule
\quad + IoP & \cmark & \xmark & \xmark & 40.72 & 31.14 \\
\quad + CL & \xmark & \xmark & \cmark & 41.43 & 31.76 \\
\midrule
\quad\quad+ TS (S) & \cmark & \cmark & \xmark & 41.54 & 31.17 \\
\quad\quad+ TS (C) & \cmark & \cmark & \xmark & 41.96 & 31.72 \\
\quad\quad+ TS (P) & \cmark & \cmark & \xmark & 41.87 & 31.67 \\
\quad\quad+ TS (P, C) & \cmark & \cmark & \xmark & 42.36 & 32.02 \\
\midrule
\quad\quad+ CL & \cmark & \xmark & \cmark & 41.77 & 32.11 \\
\rowcolor[rgb]{0.949,0.996,0.949}
\textbf{\modelname} & \cmark & \cmark & \cmark & \textbf{43.06} & \textbf{32.69} \\

\bottomrule
\end{tabular}
\caption{
Ablation study of the three innovations in \modelname. Baseline: Finetune Time-R1 with reformulated GQA data with a reward combination of tIoU, answer correctness, and format. TS(S): fixed timestamp rendering. TS(C)/(P): adaptive timestamp color/position across frames. CL: Curriculum learning.
}
\label{tab:ablation}
\vspace{-3mm}
\end{table}

\begin{table*}[t]
\centering
\small
\begin{tabular}{c c c c c c c}
\toprule
\textbf{CL Schedule} & \textbf{mIoP} & \textbf{mIoU} & \textbf{IoP@0.5} & \textbf{IoU@0.5} & \textbf{Acc@QA} & \textbf{Acc@GQA} \\
\hline

S3
& 47.44 & 31.30 & 49.36 & \textbf{29.99} & 69.31 & 37.92 \\

S3-S2
& 47.78 & 28.36 & 48.77 & 26.14 & 70.83 & 37.92 \\

S3-S2-S1
& \textbf{49.49} & \textbf{31.36} & \textbf{50.64} & 29.17 & \textbf{71.18} & \textbf{39.56} \\

\hline
S2
& 48.78 & 31.37 & 48.77 & 28.47 & 71.76 & 38.62 \\

S2-S1
& 49.56 & 31.21 & 51.11 & 30.81 & \textbf{73.75} & 40.72 \\

S2-S1-S3
& \textbf{51.16} & \textbf{33.49}
& \textbf{53.44} & \textbf{33.02}
& 70.95 & \textbf{40.96} \\

\hline
\rowcolor[rgb]{0.949,0.996,0.949}
S1-S2-S3       & \textbf{51.60} & \textbf{29.39} & \textbf{53.92} & \textbf{27.37} & \textbf{72.93} & \textbf{43.06} \\
\bottomrule
\end{tabular}
\caption{Analysis of curriculum ordering on ReXTime. The results demonstrate that the forward long-to-short ordering (S1→S2→S3) is critical, as the direction of difficulty progression (not merely stage coverage), determines grounded reasoning performance.}
\label{tab:curriculum}
\end{table*}

\begin{table*}[t!]
\centering
\small
\begin{tabular}{l ccc cccc}
\toprule
\multirow{2}{*}{\textbf{Question Type}}
& \multicolumn{3}{c}{\textbf{Causal}} 
& \multicolumn{4}{c}{\textbf{Temporal}} \\
\cmidrule(lr){2-4} \cmidrule(lr){5-8}
& \textbf{Why} & \textbf{How} & \textbf{Avg}
& \textbf{Present} & \textbf{Past} & \textbf{Future} & \textbf{Avg}\\
\midrule
TOGA \cite{gupta2025toga} 
& 26.10 & 27.40 & 26.41
& 23.40 & 18.00 & 18.10 & 20.05\\
Time-R1 \cite{wang2025time}
& 28.62 & 22.49 & 27.11
& 27.11 & 25.81 & 19.84 & 22.77\\

\rowcolor[rgb]{0.949,0.996,0.949}
\textbf{\modelname\ (Ours)} 
& \textbf{35.02} & \textbf{30.15} & \textbf{33.83}
& \textbf{32.28} & \textbf{30.11} & \textbf{27.14} & \textbf{29.17}\\
\bottomrule
\end{tabular}
\caption{Acc@GQA comparison across question types on NExT-GQA. \modelname\ consistently outperforms baselines across all question types, with the largest gain on future-oriented temporal reasoning.}
\label{tab:question_type}
\end{table*}

\subsection{Ablation Study}
Table~\ref{tab:ablation} isolates the contribution of each \modelname\ component on both benchmarks. All ablations are conducted on the Time-R1-7B backbone, with each component added incrementally to evaluate its independent and combined effect. 

The ablation results demonstrate that all three proposed components contribute consistently to GQA performance, with the full \modelname\ achieving the best results on both ReXTime and NExT-GQA. Starting from the finetuned baseline, introducing the IoP reward yields the remarkable improvement on NExT-GQA (+4\%), highlighting the importance of evidence spotting over strict temporal overlap for grounded QA.
Adding timestamp rendering (TS) further improves performance across both benchmarks, confirming the effectiveness of explicit temporal visual prompting. Among different variants, adaptive timestamp rendering consistently outperforms fixed rendering, and jointly adaptive position and color achieves the best performance, indicating that content-aware rendering better preserves timestamp visibility under diverse visual scenes.
Curriculum learning (CL) also brings additional gains, particularly on NExT-GQA, validating the effectiveness of progressively transferring reasoning capability from long to short evidence intervals. 

\begin{figure}[t!]
    \centering
    \begin{subfigure}[t]{0.48\textwidth}
        \centering
        \includegraphics[width=\linewidth]{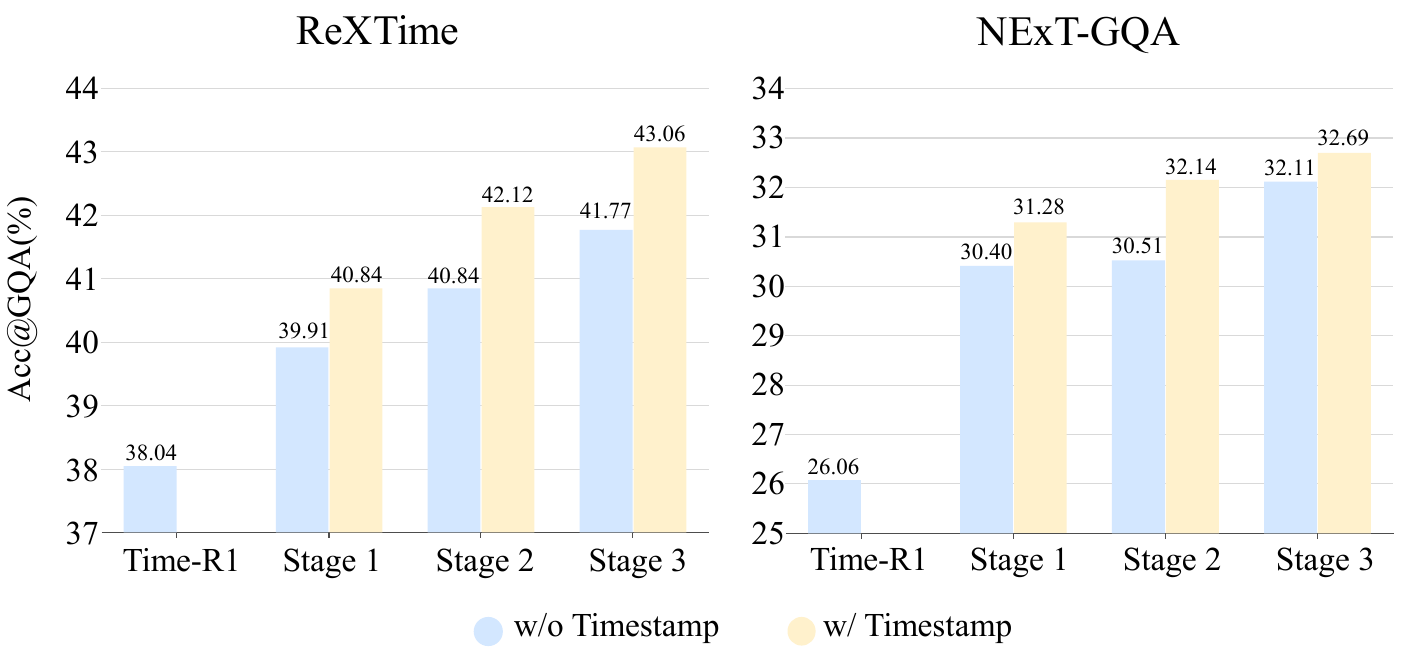}
        \caption{Time-R1 backbone.}
        \label{fig:stage_timeR1}
    \end{subfigure}
    \hfill
    \begin{subfigure}[t]{0.48\textwidth}
        \centering
        \includegraphics[width=\linewidth]{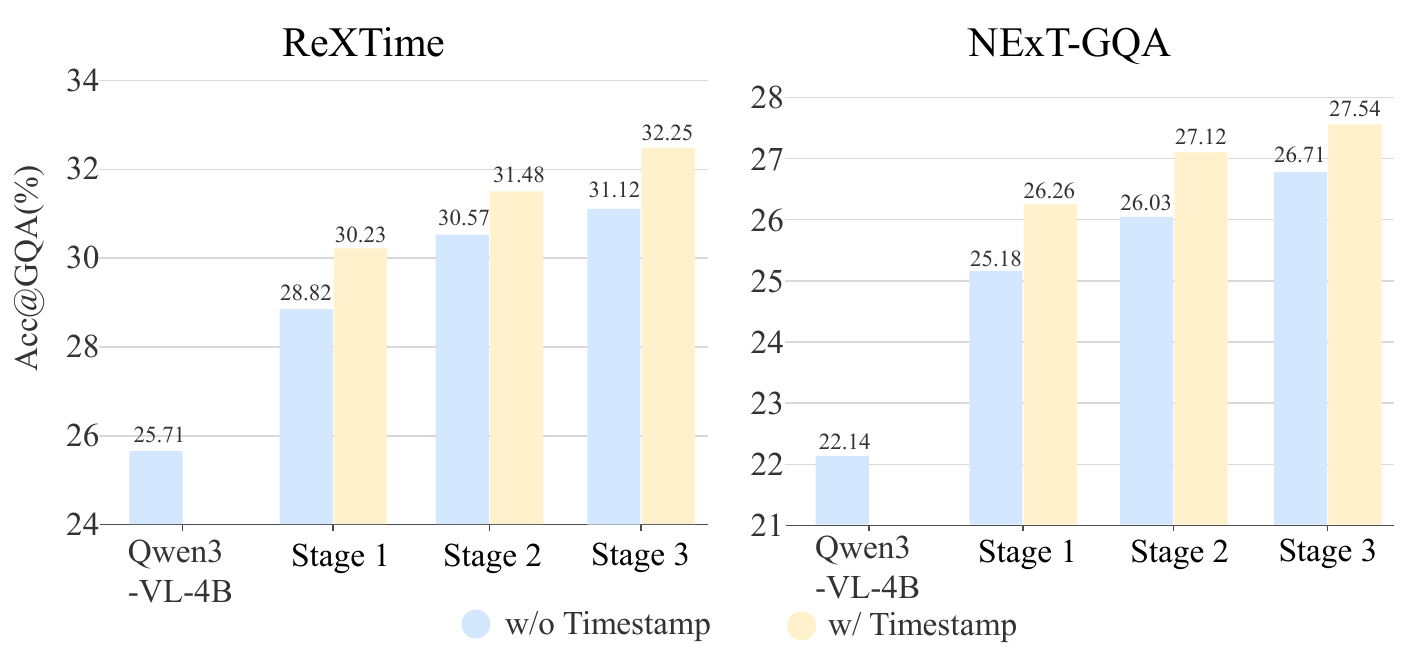}
        \caption{Qwen3-VL backbone.}
        \label{fig:stage_qwen}
    \end{subfigure}
    \caption{Acc@GQA across curriculum stages. The GQA performance consistently improves during curriculum learning, while adaptive timestamp rendering brings additional gains at each stage.}
    \label{fig:stage_compare}
    \vspace{-3mm}
\end{figure}

\subsection{Study of Curriculum Stages}

Figure~\ref{fig:stage_compare} shows the stage-wise Acc@GQA results. On both Time-R1-7B and Qwen3-VL-4B, performance improves steadily from stage 1 to stage 3, yielding over 11\% gain on ReXTime and 12\% on NExT-GQA compared with the corresponding baselines. The consistent trend across backbones demonstrates that the Long-to-Short Evidence Curriculum effectively transfers grounded reasoning ability from easier long-evidence samples to harder short-evidence ones. Adaptive timestamp rendering further provides consistent gains at every stage. In all settings, models with adaptive timestamps outperform those without rendering, indicating that explicit temporal cues complement curriculum-based training and jointly improve evidence grounding.

\subsection{Reward-based vs. Evidence-based Curriculum}

\begin{table}[t]
\centering
\resizebox{\linewidth}{!}{
\begin{tabular}{lccccc}
\toprule
\multicolumn{6}{c}{\textbf{NExT-GQA}}\\
\midrule
Method & mIoP & mIoU & IoP@0.5 & IoU@0.5 & Acc@GQA \\
baseline & 36.0 & 30.7 & 31.9 & 25.3 & 26.1 \\
Rewards & 41.2 & 33.4 & 39.0 & 29.8 & 30.5 \\
evidence duration & 42.7 & 33.5 & 41.1 & 29.8 & 32.7 \\
\midrule
\multicolumn{6}{c}{\textbf{ReXTime}}\\
\midrule
Method & mIoU & IoU@0.3 & IoU@0.5 & Acc & Acc@GQA \\
baseline & 28.40 & 39.53 & 27.37 & 71.18 & 21.46 \\
Rewards & 28.13 & 41.28 & 25.01 & 69.25 & 20.36 \\
evidence duration & 29.39 & 41.64 & 27.37 & 72.93 & 21.87 \\
\bottomrule
\end{tabular}}
\caption{An effectiveness comparison experiment between curriculum learning based on rewards and curriculum learning based on evidence duration.}
\label{tab:reward_curriculum}
\end{table}

\begin{table}[t]
\centering
\small
\begin{tabular}{lcccc}
\toprule
\textbf{Method} & \textbf{Time-R1} & \textbf{Stage 1} & \textbf{Stage 2} & \textbf{Stage 3} \\
\midrule
Relative & 26.06 & 31.55 & 32.03 & 32.17 \\
Absolute & 26.06 & 31.28 & 32.14 & \textbf{32.69} \\
\bottomrule
\end{tabular}
\caption{Comparison of curriculum learning based on absolute and relative evidence duration in terms of Acc@GQA on NExT-GQA.}
\label{tab:relative_curriculum}
\end{table}

We implement a reward-based curriculum using the same GQA rewards as LOVER (IoP, format, and choice rewards). We rank samples by zero-shot reward scores and train progressively from high- to low-reward samples. Table~\ref{tab:reward_curriculum} shows this strategy improves IoP-related performance on NExT-GQA, but it underperforms our evidence-duration curriculum and degrades IoU-related metrics on ReXTime, indicating its sensitivity to reward design and potential optimization bias. In contrast, our curriculum relies solely on evidence duration as an intrinsic difficulty measure, making it more robust and model-agnostic.

\subsection{Absolute vs. Relative Evidence Duration}

\begin{figure*}[t!]
    \centering
    \includegraphics[width=\textwidth]{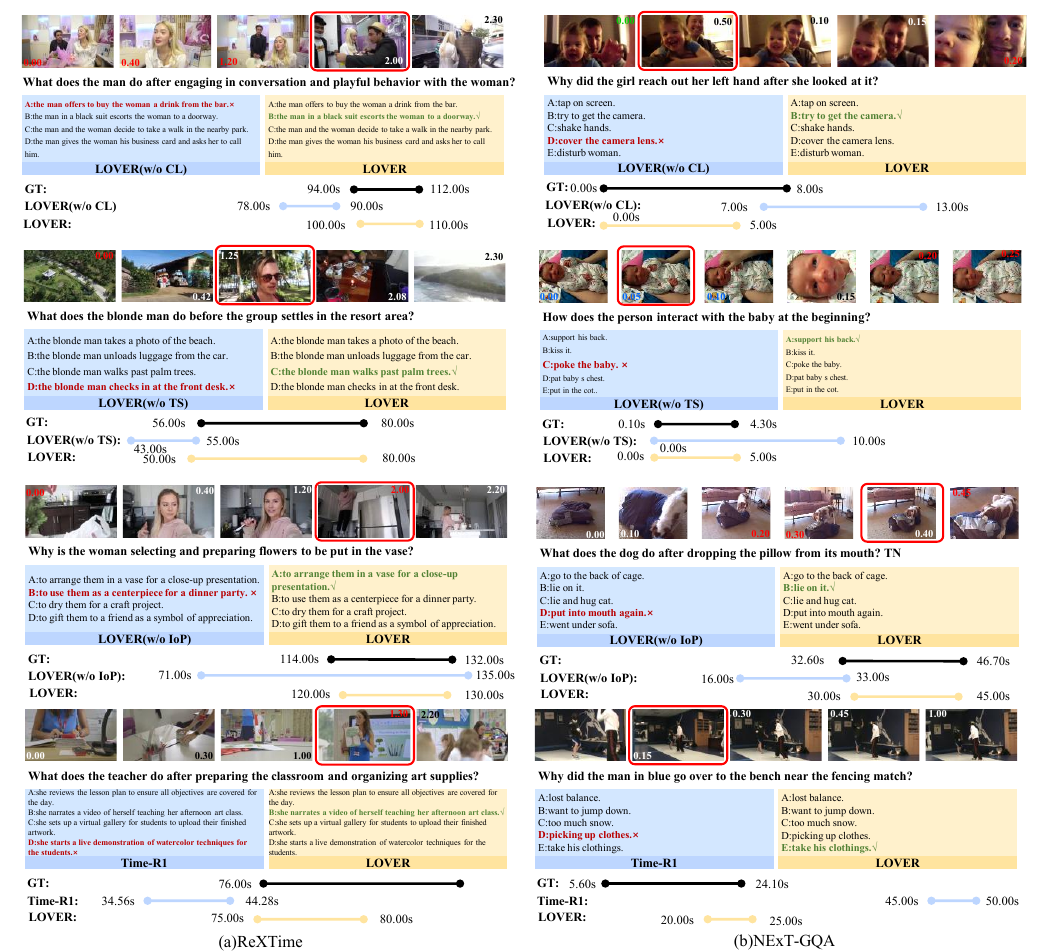}
    \caption{GQA result visualization. The timestamps overlaid on video frames are formatted as MM.SS, while the temporal intervals displayed below are in seconds. \modelname\ demonstrates stronger evidence discovery capability than its ablated variants and Time-R1, producing temporal predictions that more accurately cover the ground-truth evidence regions and thereby brings more accurate answer.}
    \label{fig:qualitative_main}
\end{figure*}

In order to explore the scheme for defining the difficulty of evidence-based curriculum learning, we compared two curriculum strategies based on the duration of relative evidence and absolute evidence. The relative strategy was based on the proportion of the evidence duration to the total video duration of the sample, while the absolute strategy directly used the duration of the evidence segments as the difficulty measurement indicator.

Table~\ref{tab:relative_curriculum} shows that during the curriculum training process, both strategies continuously improved performance based on the Time-R1 baseline. However, the relative strategy improved by 6.11\% after three stages, while the absolute strategy improved by 6.63\%. In contrast, the absolute evidence duration was more effective in estimating the difficulty of curriculum learning.

\subsection{Effect of Curriculum Ordering}
Table~\ref{tab:curriculum} analyzes the effect of curriculum ordering on ReXTime. Among all evaluated schedules, the forward ordering S1$\to$S2$\to$S3 — progressing from long to short evidence — consistently achieves the best performance across both grounding and grounded QA metrics. This result is not merely attributable to data coverage: the reverse schedule S3$\to$S2$\to$S1 exposes the model to the same three stages yet performs substantially worse, demonstrating that the \textit{ordering} of difficulty, rather than stage inclusion alone, is the critical factor.

Training exclusively on short-evidence samples (S3) yields notably inferior grounded reasoning, suggesting that the model fails to establish the coarse temporal alignment necessary for reliable evidence localization when hard samples are introduced without prerequisite exposure to easier ones. Schedules that begin from medium-evidence samples (S2-first) show intermediate performance, further supporting the view that grounded reasoning ability is most stably acquired through a gradual transition from dense, redundant evidence to sparse, precisely localized evidence. Taken together, these results validate the long-to-short design principle underlying \modelname's curriculum: effective sparse-evidence reasoning depends on first building coarse alignment on long-evidence samples before adapting to finer-grained temporal localization.

\section{Question-Type Analysis}
Table~\ref{tab:question_type} reports Acc@GQA across question categories on NExT-GQA. \modelname\ significantly improves over Time-R1 across all question types, with the largest gain in \textit{Future} questions (\texttt{``what happens next?''}) (+7.3\%, from 19.8\% to 27.1\%). Future reasoning demands strong causal reasoning capability where precise evidence grounding plays pivotal role. This conclusion is also consolidated by the consistent gains in causal categories (\texttt{`Why', `How'}). Thus, the strong improvements here have validated the strength of \modelname.

\subsection{Qualitative Analysis}
Figure~\ref{fig:qualitative_main} presents qualitative comparisons across ablation configurations on ReXTime and NExT-GQA. Without IoP reward (The 1st row), the model produces temporal predictions that deviate from the ground-truth evidence region, while the full \modelname\ recovers the correct evidence span and answers correctly. Without curriculum learning (The 2nd row), the model anchors to irrelevant segments, whereas \modelname\ with curriculum training identifies the correct evidence region across varying temporal difficulty. Removing timestamp rendering (The 3rd row) leads to inaccurate boundary prediction despite visually identifiable evidence, confirming that explicit temporal cues contribute to more reliable evidence spotting. Finally, compared to Time-R1, \modelname\ consistently discovers the supporting evidence in both long- and short-evidence cases, validating the overall effectiveness of the proposed framework.


%% file: secs/conclusion.tex
\section{Conclusion}

We present \textbf{\modelname}, an RL-based post-training framework for grounded video QA that addresses the imbalanced difficulty of temporal evidence and the misalignment between existing reward objectives and GQA requirements. Through three architecture-agnostic components, namely long-to-short evidence curriculum, GQA-specific IoP rewards, and adaptive timestamp rendering, \modelname\ consistently improves grounded reasoning across backbones and benchmarks, with the most pronounced gains on short-evidence samples where existing methods fall short. With these efforts, we hope our designs and superior results help advance grounded reasoning for trustable MLLMs.

\section*{Limitations}
\modelname\ currently focuses primarily on temporal grounding over visual frames while overlooking auxiliary modalities such as audio and subtitles. Although visual evidence is sufficient for many grounding scenarios, some questions intrinsically rely on speech content, environmental sounds, or textual cues that cannot be fully inferred from frames alone. Extending \modelname\ toward multimodal evidence grounding may further improve grounded reasoning performance, particularly for audio-dependent causal and temporal questions.

Moreover, the current framework is evaluated mainly under multiple-choice GQA setting. Its effectiveness for open-ended GQA remains unexplored. Future work may investigate how to extend evidence-aware reward design from discrete answer selection to generative answer evaluation, enabling more flexible grounded reasoning and free-form response generation.

\section*{Acknowledgments}
This research is partly supported by the advanced computing resources provided by the Supercomputing Center of USTC.

%% file: secs/appendix.tex
\section{Prompt Templates}

We provide the prompt templates used for training data construction and inference to ensure full reproducibility of \modelname.

\subsection{Question-Answer Pair Generation}
\begin{figure*}[t]
    \centering
    \includegraphics[width=0.9\linewidth]{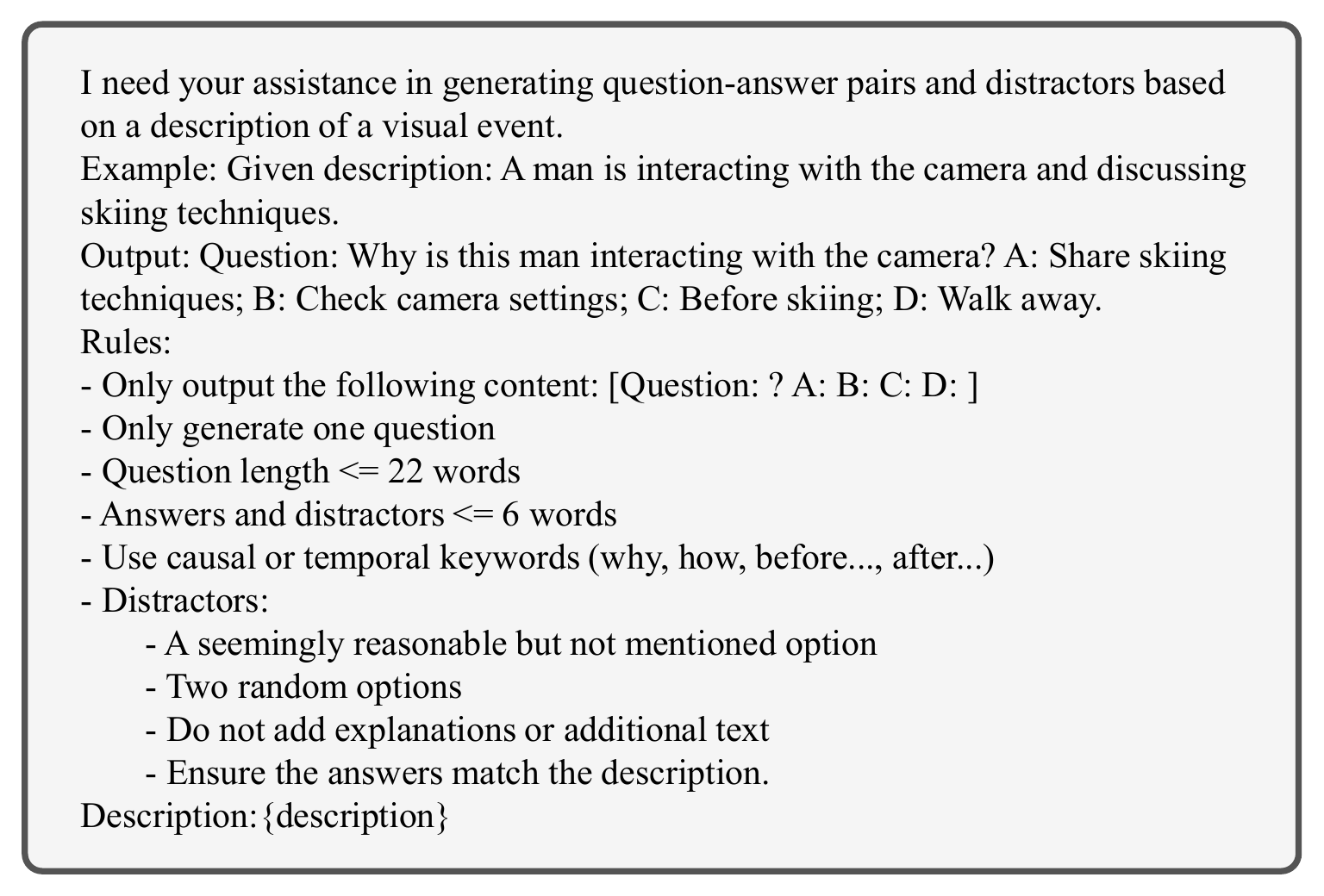}
    \caption{Prompt template used to generate grounded question–answer pairs from video descriptions.}
    \label{fig:prompt_gen}
\end{figure*}

Figure~\ref{fig:prompt_gen} shows the prompt template for reformatting Time-R1 training data into GQA-style supervision via DeepSeek. The template enforces causal or temporal keywords and structured multiple-choice formatting.

\subsection{Inference Prompt}
\begin{figure*}[t]
    \centering
    \includegraphics[width=0.9\linewidth]{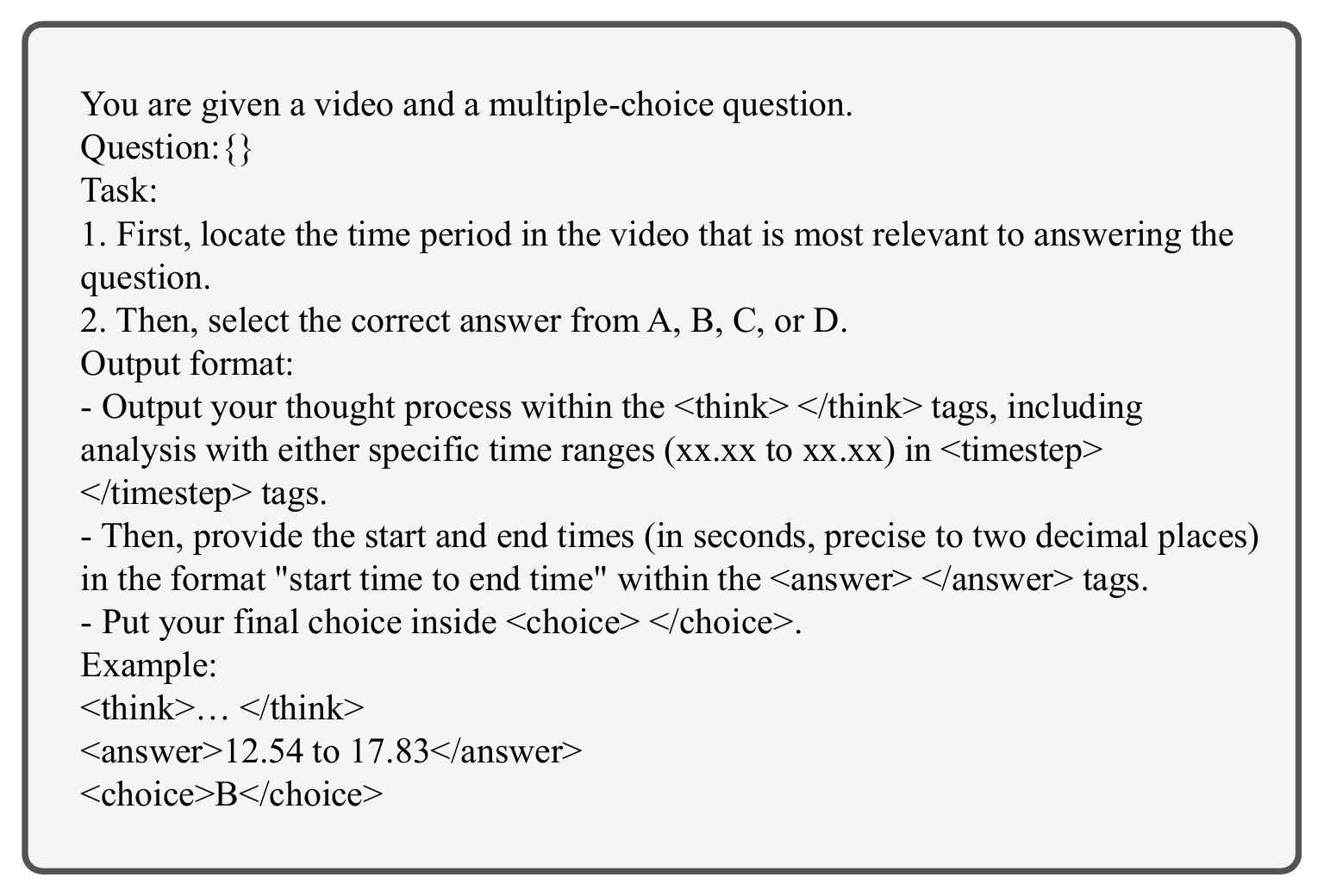}
    \caption{Prompt template for grounded video question answering without timestamp rendering. This figure presents the inference prompt template that instructs the model to output its reasoning process, predicted temporal interval, and final answer choice in structured tags.}
    \label{fig:prompt_no_ts}
\end{figure*}

\begin{figure*}[t]
    \centering
    \includegraphics[width=0.9\linewidth]{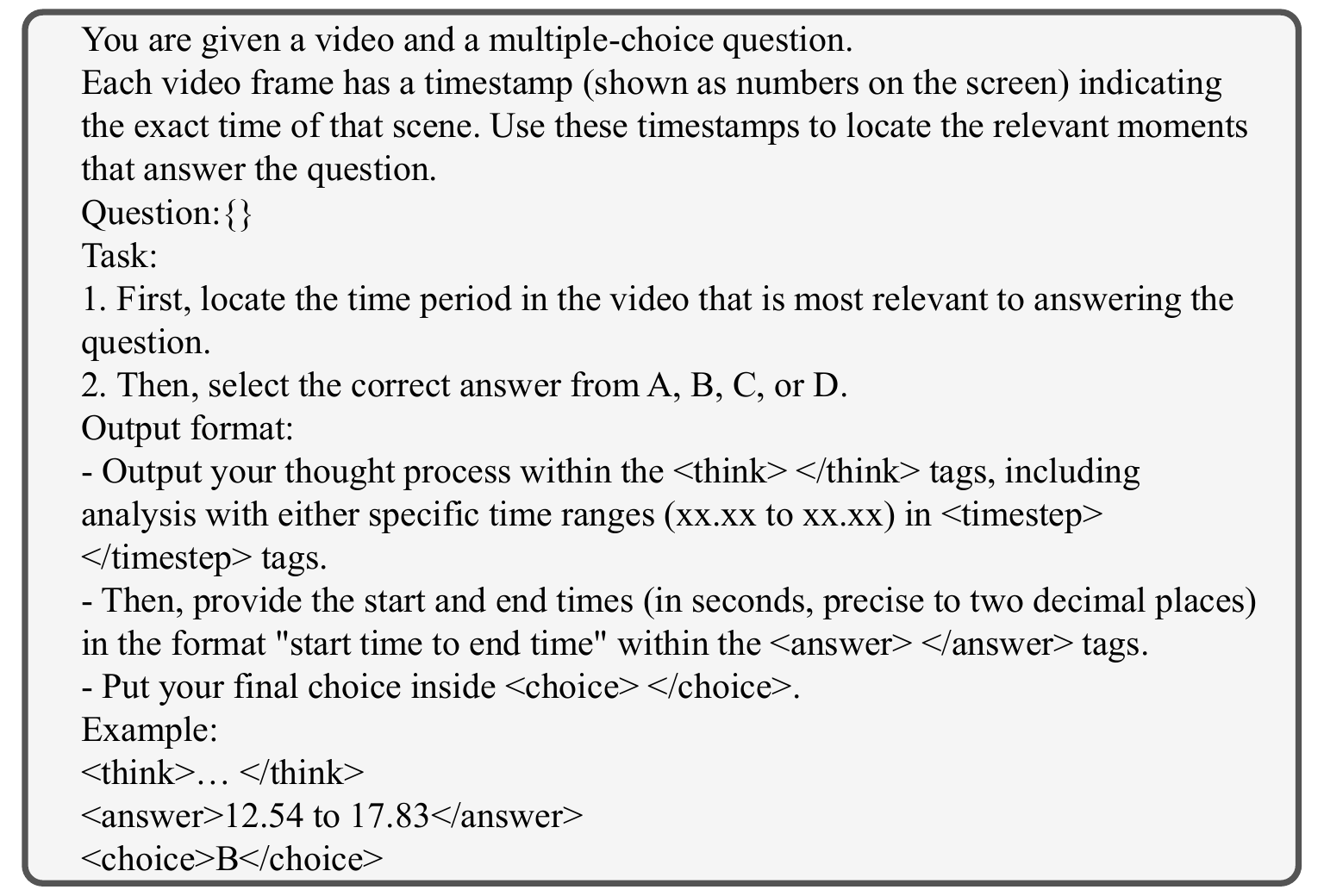}
    \caption{Prompt template for grounded video question answering with adaptive timestamp rendering. This figure presents the inference prompt template used when frames carry rendered timestamps, additionally instructing the model to reference the visible frame-level time cues when localizing the relevant video segment.}
    \label{fig:prompt_ts}
\end{figure*}

Figures~\ref{fig:prompt_no_ts} and~\ref{fig:prompt_ts} show the inference templates without and with adaptive timestamp rendering, respectively. Both require the model to output reasoning in \texttt{<think>} tags, the predicted interval in \texttt{<answer>} tags, and the final choice in \texttt{<choice>} tags. The timestamp-enabled template additionally instructs the model to exploit rendered frame-level time references during grounded reasoning.

\section{Data Quality Verification}
\label{Quality_Verification}

Since our training data is constructed by converting textual descriptions into grounded question-answer pairs using LLMs, we further evaluate the quality of the generated annotations. We randomly sampled 500 QA instances (20\% of the dataset) and manually inspected whether the questions require video understanding, whether the distractors are reasonable, and whether the annotated temporal intervals are correct.

The inspection process took approximately 10 hours and achieved an acceptance rate of 96.8\%. This high-quality ratio is mainly attributed to the careful filtering process of the original Time-R1 dataset, where the source video descriptions were already manually curated. In our data construction pipeline, the LLM is only used to transform existing textual descriptions into GQA-style questions rather than generating new video content.

Furthermore, \modelname\ consistently improves performance despite the remaining noise in automatically generated annotations, demonstrating the robustness of our evidence-based curriculum learning framework under imperfect supervision.

\section{Implementation Details of Adaptive Timestamp Rendering}
This section describes the implementation of the adaptive timestamp rendering algorithm, which processes each sampled video frame independently through four sequential steps.

\noindent\textbf{Step 1: Background Region Detection.}
Each video frame is first segmented into three color clusters via K-Means clustering on pixel RGB values. The cluster with the largest pixel count is treated as the dominant background. Connected component analysis is then applied to this cluster mask, and the largest connected component is retained as the primary background region, filtering out small isolated patches.

\begin{table}[t]
\centering
\small
\setlength{\tabcolsep}{6pt}
\renewcommand{\arraystretch}{1.08}
\begin{tabular}{lcc}
\toprule
\textbf{Models} 

& \textbf{Acc@QA} 
& \textbf{Acc@GQA} \\
\midrule
\multicolumn{3}{l}{\cellcolor[rgb]{0.898,0.898,0.898}\textit{REXTIME}} \\
Qwen3-VL-4B & 65.19 & 25.71 \\
Stage 1 & 65.68 & 28.82 \\
\rowcolor[rgb]{0.949,0.996,0.949}
\textbf{Stage 1} 
&  65.61
&  30.23\\
Stage 2 & \textbf{65.72}  & 30.57 \\
\rowcolor[rgb]{0.949,0.996,0.949}
\textbf{Stage 2} 
& 65.03
& 31.48\\
Stage 3 & 65.57 & 31.12 \\
\rowcolor[rgb]{0.949,0.996,0.949}
\textbf{Stage 3} 
& 65.49
& \textbf{32.35} \\
\midrule
\multicolumn{3}{l}{\cellcolor[rgb]{0.898,0.898,0.898}\textit{NExT-GQA}} \\
Qwen3-VL-4B & 68.45 & 22.14 \\
Stage 1 & \textbf{68.91}  & 25.18 \\
\rowcolor[rgb]{0.949,0.996,0.949}
\textbf{Stage 1} 
& 68.36
& 26.26 \\
Stage 2 & 68.13 & 26.03 \\
\rowcolor[rgb]{0.949,0.996,0.949}
\textbf{Stage 2} 
& 68.14
& 27.12 \\
Stage 3 & 68.63 & 26.71 \\
\rowcolor[rgb]{0.949,0.996,0.949}
\textbf{Stage 3} 
& 68.59
& \textbf{27.54} \\
\bottomrule
\end{tabular}
\caption{Stage-wise Acc@QA and Acc@GQA of Qwen3-VL-4B on REXTIME and NExT-GQA across curriculum stages. \colorbox{green!15}{Shaded rows} indicate settings with adaptive timestamp rendering; plain rows are without timestamp rendering.This table demonstrates that Acc@GQA improves monotonically across stages while Acc@QA remains stable, confirming that curriculum training enhances temporal localization without compromising answer generation accuracy.}
\label{tab:qwen_stages}
\end{table}

\begin{table*}[t]
\centering
\resizebox{\textwidth}{!}{
\begin{tabular}{lcccccc}
\toprule
Dataset & Run & mIoP & mIoU & IoP@0.5 & IoU@0.5 & Acc@GQA \\
\midrule
\multirow{4}{*}{NExT-GQA}
& Seed 1 & 42.7 & 33.5 & 41.1 & 29.7 & 32.7 \\
& Seed 2 & 42.8 & 33.7 & 41.3 & 29.9 & 32.8 \\
& Seed 3 & 42.7 & 33.4 & 41.1 & 29.8 & 32.7 \\
& Mean$\pm$Std & $\mathbf{42.7\pm0.0}$ & $\mathbf{33.5\pm0.1}$ & $\mathbf{41.2\pm0.1}$ & $\mathbf{29.8\pm0.0}$ & $\mathbf{32.7\pm0.0}$ \\
\midrule
\multicolumn{7}{c}{}\\[-1.2ex]
\toprule
Dataset & Run & mIoU & IoU@0.3 & IoU@0.5 & Acc & Acc@GQA \\
\midrule
\multirow{4}{*}{ReXTime}
& Seed 1 & 29.39 & 41.64 & 27.37 & 72.93 & 21.87 \\
& Seed 2 & 29.18 & 41.26 & 27.15 & 72.87 & 21.73 \\
& Seed 3 & 29.41 & 41.61 & 27.42 & 73.03 & 21.89 \\
& Mean$\pm$Std & $\mathbf{29.33\pm0.10}$ & $\mathbf{41.50\pm0.17}$ & $\mathbf{27.31\pm0.12}$ & $\mathbf{72.94\pm0.07}$ & $\mathbf{21.83\pm0.07}$ \\
\bottomrule
\end{tabular}}
\caption{Performance variance of \modelname\ under different random seeds on NExT-GQA and ReXTime. Mean and standard deviation are reported over three independent runs.}
\label{tab:seed}
\end{table*}

\begin{table}[t]
\centering
\resizebox{\linewidth}{!}{
\begin{tabular}{lccc}
\toprule
\textbf{Model} & \textbf{mIoU} $\uparrow$ & \textbf{Rec.@IoU} $\uparrow$ & \textbf{Acc.@IoU} $\uparrow$ \\
\midrule
Human & 35.5 & 51.2 & 29.8 \\
\midrule
Video-LLAVA (7B) \cite{li2024llava}        & 1.13 & 1.96 & 0.59 \\
VideoLLAMA (7B) \cite{zhang2023video}         & 1.21 & 1.87 & 0.84 \\
VideoChat2 (7B)  \cite{li2025videochat}        & 1.28 & 1.98 & 0.94 \\
ST-LLM (7B) \cite{liu2024st}             & 2.23 & 2.86 & 1.13 \\
ViLA (8B)   \cite{lin2024vila}             & 1.56 & 2.89 & 1.35 \\
MiniCPM-v2.6 (8B) \cite{yao2024minicpm}       & 2.35 & 2.61 & 1.04 \\
LongVA (7B) \cite{zhang2024long}             & 2.94 & 3.86 & 1.78 \\
LLaVA-OneVision (7B) \cite{li2024llava}    & 1.63 & 1.78 & 1.08 \\
Kangaroo (8B) \cite{liu2026kangaroo}     & 2.56 & 2.81 & 1.94 \\
Time-R1(7B) \cite{wang2025time}   & 2.68 & 3.14 & 1.47 \\
\rowcolor[rgb]{0.949,0.996,0.949}
\textbf{LOVER(7B)}            & \textbf{3.16} & \textbf{3.88} & \textbf{2.23} \\
\bottomrule
\end{tabular}
}
\caption{Results on CG-Bench benchmarks. mIoU denotes the mean temporal IoU over all samples; Rec.@IoU is the average recall under IoU thresholds from 0.1 to 0.5; Acc.@IoU requires both a correct answer and a temporal localization IoU above the corresponding threshold.}
\label{tab:cgbench}
\end{table}

\noindent\textbf{Step 2: Corner Position Selection.}
For each of the four frame corners (top-left, top-right, bottom-left, bottom-right), we identify the background pixel geometrically closest to that corner. The corner whose nearest background pixel lies closest to the actual frame corner is selected as the timestamp placement location, ensuring the timestamp occupies a corner-aligned background region with minimal foreground occlusion.

\noindent\textbf{Step 3: Adaptive Font Color Selection.}
The font color is chosen from a fixed candidate set {black, white, red, green, blue} based on perceptual luminance contrast. The brightness of the background pixel at the selected position is computed via the standard luminance formula, and the candidate color with the maximum absolute brightness difference is selected, ensuring readability against varying background appearances.

\noindent\textbf{Step 4: Timestamp Overlay.}
The real-world timestamp for each frame is computed from the segment start time and sampled frame rate, formatted as MM.SS, and rendered onto the frame using anti-aliased text. The rendered frame is subsequently passed to the visual encoder at the spatially resized resolution consumed by the model.

\section{Stage-wise Results for Qwen3-VL-4B}

Table~\ref{tab:qwen_stages} supplements the main paper results by reporting Acc@QA alongside Acc@GQA for Qwen3-VL-4B across curriculum stages, providing a more complete picture of how grounding and answering evolve jointly during training.

The key observation is that Acc@GQA improves monotonically across all three curriculum stages on both benchmarks, with and without adaptive timestamp
rendering, mirroring the trend reported for Time-R1-7B in Figures~\ref{fig:stage_timeR1} and~\ref{fig:stage_qwen}. This consistency
across backbones of different scales confirms the architecture-agnostic nature of the Long-to-Short Evidence Curriculum. Adaptive timestamp rendering
provides additional gains at every stage, further corroborating the complementarity between curriculum training and explicit temporal cues
established in the main ablation study (Table~\ref{tab:ablation}). 

Crucially, Acc@QA remains largely stable throughout training, indicating that the observed Acc@GQA improvements stem from more accurate temporal
localization rather than shifts in answer generation. This decoupling supports the central claim of \modelname: that evidence-aware curriculum
design and IoP-based reward optimization improve grounded reasoning without compromising QA accuracy.

\section{Evaluation on CG-Bench}
\label{CG-Bench}

To further evaluate the generalization ability of \modelname\ on long-video understanding, we extend our experiments to CG-Bench, a challenging clue-grounded question answering benchmark. The benchmark provides 12,129 question-answer pairs across perception, reasoning, and hallucination evaluation settings. Notably, all videos are longer than 30 minutes, making CG-Bench a challenging testbed for long-context video reasoning.

As shown in Table~\ref{tab:cgbench}, \modelname\ achieves the best performance among all compared 7B/8B open-source models. Specifically, \modelname\ obtains 3.16 mIoU, 3.88 Rec.@IoU, and 2.23 Acc.@IoU, outperforming previous strong baselines such as Time-R1 and LongVA. These results demonstrate that our evidence-aware curriculum learning and temporal grounding strategy generalize effectively to challenging long-video scenarios.

\section{Performance Variance under Different Random Seeds}

To evaluate the stability and reproducibility of \modelname\, we conduct experiments with three different random seeds on NExT-GQA and ReXTime. We report the mean and standard deviation of all evaluation metrics across independent runs.

As shown in Table~\ref{tab:seed}, the variances are very small (e.g., 32.7 ± 0.0 on NExT-GQA and 21.83 ± 0.07 on ReXTime for Acc@GQA), demonstrating that \modelname\ is stable and reproducible. The minor differences on NExT-GQA are due to rounding to one decimal place, indicating that random seeds have a negligible impact on overall performance.

